%% file: main.tex
\documentclass[letterpaper, 10 pt, conference]{ieeeconf}  

\IEEEoverridecommandlockouts                              

\makeatletter
\def\fnum@table{Table~\thetable}
\makeatother
\input{our_header.tex}

\title{\LARGE \bf
Extending Ground-Constraint LiDAR-IMU Calibration to Tilted Surfaces
in a Continuous-Time Framework
}

\author{Vassili Korotkine$^{1}$, Pierre Chamoun$^{1}$, Mohammed Ayman Shalaby$^{2}$, and James Richard Forbes$^{1}$
\thanks{This work was funded via the Natural Sciences and Engineering Research Council of Canada (NSERC)
Alliance Grant in collaboration with Denso Corporation, as well as Rheinmetall Canada.}
\thanks{$^{1}$The authors are with the Department of Mechanical Engineering, McGill
University, Montreal, Canada (e-mails:
\texttt{\small{vassili.korotkine@mail.mcgill.ca}},
\texttt{\small{pierre.chamoun@mail.mcgill.ca}},
\texttt{\small{james.richard.forbes@mcgill.ca}}).}
\thanks{$^{2}$The author is with Rheinmetall Provectus, Ottawa, ON, Canada
(e-mail: \texttt{\small{mohammed.shalaby@mail.mcgill.ca}}).}%
}

\begin{document}

\maketitle
\thispagestyle{empty}
\pagestyle{empty}

\begin{abstract}
This paper presents a novel method that extends
targetless LiDAR-IMU calibration for ground vehicles to non-flat environments. 
Calibration typically necessitates full excitation of the sensor rig, 
a requirement that is not fulfilled by ground vehicles in normal operation. 
To address the degenerate planar motion, state-of-the-art methods propose residuals that assume the colinearity of the
gravity and physical surface normal vectors, restricting usage to cases where the ground is assumed
flat. 
This paper proposes ground-plane residuals that do not require this assumption, and are applicable for planar motion on a
tilted surface.
Results are demonstrated on a dataset collected from a Husky ground vehicle, on the M2DGR dataset, as well as on an offroad vehicle
dataset.
Repeatability is shown to be improved both in tilted and flat-ground scenarios, with strong improvement demonstrated
for the tilted case. 
The implementation and experiments are open-sourced at \url{https://anonymous.4open.science/r/licalib_tilted_ground-9885}.
\end{abstract}
\input{sections/introduction.tex}

\input{sections/proposed_method.tex}
\input{sections/results.tex}
\input{sections/conclusion.tex}
\FloatBarrier
\input{sections/acknowledgements.tex}

\printbibliography

\addtolength{\textheight}{-12cm}   

\end{document}

%% file: our_header.tex
\newcommand{\mbs}[1]{\ensuremath{\boldsymbol{#1}}}
\DeclareMathAlphabet{\mbf}{OT1}{ptm}{b}{n}
\newcommand{\mc}[1]{\ensuremath{\mathcal{#1}}}

\usepackage{xcolor}
\usepackage{amsmath}
\usepackage{amssymb}
\usepackage{mathtools}
\usepackage[backend=bibtex,bibstyle=ieee,citestyle=numeric-comp,doi=false,isbn=false]{biblatex}
\usepackage{xspace}
\usepackage{multirow}
\usepackage{booktabs}
\usepackage{placeins}

\newcommand{\trans}{{\ensuremath{\mathsf{T}}}}
\newcommand{\mbfbar}[1]{\ensuremath{\bar{\mbf{#1}}}}

\newcommand{\mbftilde}[1]{\ensuremath{\tilde{\mbf{#1}}}}

\DeclareMathOperator*{\argmin}{arg\,min}

\newcommand{\norm}[1]{\ensuremath{\left\Vert#1\right\Vert}}

\newcommand{\dxi}{\delta \mbs{\xi}}
\newcommand{\dxibar}{\delta \bar{\mbs{\xi}}}

\newcommand{\Log}{\text{Log}}

\newcounter{daggerfootnote}

\newcommand{\rframe}[1]{{\ensuremath{\mathcal{F}}_{#1}}}
\newcommand{\ura}[1]{{\underrightarrow{{#1}}}}

\newcommand{\NumCtrlPoints}{{N}}
\newcommand{\CtrlPtIdx}{{i}}

\newcommand{\FullState}{{\mc{X}_{\text{full}}}}

\newcommand{\NumResults}{{n_{\text{res}}}}
\newcommand{\IdxResults}{i}

%% file: sections/introduction.tex
\section{Introduction}
Localization from sensor data is a central task for robot operation. 
Multiple sensors are commonly used to provide redundancy, as well as to complement
each other. In particular, Light Detection And Ranging (LiDAR) and Inertial Measurment Unit (IMU) sensors are commonly used together, where the IMU is used for motion prediction as well as
to account for LiDAR point cloud distortion~\cite{xu2021fastlio,xu2022fastlio2,vizzo2023kiss}.
However,
a prerequisite for fusing information from the two sensors is accurate
\emph{sensor calibration}, which provides the ability to resolve data from the
different sensors in the same reference frame and relative to the same clock.
In particular, the extrinsic pose transformation (rotation and translation)
as well as time offset between the two sensors must be known. 
Inaccurate sensor calibration can cause downstream issues in a robotics pipeline, making reliable
and accurate calibration important for robot operation~\cite{li2023gnss}. 
%
\par
General calibration methods may be split into target-based and targetless approaches. 
Target-based approaches, while accurate, necessitate custom targets that
may not be readily available during deployment, and
can
be expensive to manufacture.
On the other hand, targetless approaches~\cite{lv2020targetless, zhu2022liinit,oaLv2022} do not
require any prior setup for operation, but may be less accurate due to the lack of geometric information
provided by a known target. 
Recent LiDAR-IMU calibration methods simultaneously solve for the IMU trajectory and calibration parameters
using a nonlinear-least-squares solver.
The IMU trajectory may be parametrized as a B-Spline~\cite{lv2020targetless,oaLv2022,gctHe2025,slope2025xiao}, a Gaussian process~\cite{li2021structured},
or as a sequence of discrete poses~\cite{zhu2022liinit,gril2024kim}. 
Together, the B-Spline and Gaussian process parametrizations are termed continuous-time methods,
while the discrete sequence parametrization is termed a discrete-time method. 
Existing discrete-time methods~\cite{zhu2022liinit, gril2024kim}
introduce discrete robot states at the LiDAR timestamps, and linearly interpolate the IMU data
to match the lower-frequency LiDAR stamps.
The continuous-time methods are advantageous in that they naturally allow for use
of all IMU measurements. 
%
\begin{table}[t]
\centering
\scriptsize
\caption{Comparison of targetless LiDAR--IMU calibration methods.}
\begin{tabular}{lccc}
\toprule
Method & Continuous-Time & Ground Constraint & Open Source \\
\midrule
LI-Init~\cite{zhu2022liinit} & No  & No               & Yes \\
OA-Calib~\cite{oaLv2022}     & Yes & No               & Yes \\
GRIL-Calib~\cite{gril2024kim}& No  & Flat only        & Yes \\
Slope-Based~\cite{slope2025xiao}
                             & No  & No      & No \\
GCT-LICalib~\cite{gctHe2025} & Yes & Flat only        & No \\
Proposed                     & Yes & Tilted surfaces  & Yes \\
\bottomrule
\end{tabular}
\label{tab:related_work}
\end{table}
\par
Generally, calibration methods require full motion excitation of the sensor rig
to ensure observability of the extrinsic calibration parameters~\cite{chen2025ikalibr}. 
This is an issue for ground vehicle sensor calibration, since ground vehicles commonly
undergo the single-axis rotation motion profile, which causes the position calibration extrinsic to become unobservable
along the axis of rotation~\cite{slope2025xiao, lee2024degenerate, yang2023online}.
The ground vehicle motion degeneracy has prompted research into ground constraints that can be added into the problem to
ensure observability. 
\par
%
Existing methods for ground-vehicle LiDAR-IMU calibration, GRIL-Calib~\cite{gril2024kim} and GCT-Calib~\cite{gctHe2025},
extract the ground plane from the LiDAR point cloud, yielding a normal vector and distance from the ground. 
Colinearity of the gravity and surface normal vector is then assumed to create a constraint on the transformation between
the LiDAR and the IMU. However, in the case of a tilted surface, the gravity is no longer colinear with the
physical surface normal and this assumption is no longer valid, limiting their applicability to flat ground.
Furthermore, the ground-constraint residuals of~\cite{gril2024kim,gctHe2025} overparametrize the
state by introducing ground reference frames that are unique only up to a yaw transformation.
While this works in practice, it adds unnecessary interpretation and implementation complexity.  
%
%
The slope-based method of~\cite{slope2025xiao} proposes an observability analysis that shows
that the unobservabilities disappear with pitch excitation of the vehicle, and proposes pitch excitation
as a solution to restore calibration parameter observability. 
However, the requirement for pitch excitation does not address the case of strictly planar motion. 
Table~\ref{tab:related_work}
presents an overview of LiDAR-IMU calibration systems and their relevant characteristics.
\par
In summary, existing targetless LiDAR-IMU calibration systems either require full motion excitation or
rely on a flat-ground assumption for introduction of ground constraints. 
To address these limitations, this work proposes
ground-constraint residuals that are independent of the flat ground assumption. 
A distance residual is formulated by 
comparing the LiDAR-IMU height difference to the position extrinsic projection onto the surface normal,
while an inclination residual is formed as the dot product between the surface normal and the gravity vector. 
The distance residual directly constrains the position extrinsic along the surface normal,
addressing the unobservability caused by planar motion. The inclination residual
extends the orientation ground-constraint residual of~\cite{gril2024kim} to the tilted surface case.

\par
To the best of the author's knowledge, this is the first work that addresses
the planar motion observability loss while remaining valid for tilted terrain.
Furthemore, while the proposed work is not the first continuous-time ground-constraint system,
it contributes the first open-source implementation
of a continuous-time ground-constraint system for the community to utilize and validate against. 
\par
The contributions of this paper thus consist of
\begin{enumerate}
    \item a ground-plane constraint formulation, consisting of a distance and inclination residual,
    that allows for LiDAR-IMU spatial-temporal calibration during single-axis degenerate motion \emph{without assuming that the ground is flat},
    \item a simplified formulation of the flat-ground orientation residual that does not require introduction of extraneous
    reference frames, and
    \item an open-source implementation, built within the framework of~\cite{oaLv2022}, available at \url{https://anonymous.4open.science/r/licalib_tilted_ground-9885}.
\end{enumerate}
\par 
The proposed method is compared to GRIL-Calib~\cite{gril2024kim}, an open-source
ground-constraint-based LiDAR-IMU calibration system.
The proposed method is evaluated on
self-collected flat-ground and inclined-ground datasets
from two different vehicle platforms, 
as well as on the openly available M2DGR dataset~\cite{yin2022m2dgr}. 
The self-collected datasets are openly available at~\cite{anonymous_2026}.
The inclined-ground sequences are used to demonstrate
degraded performance for the baseline while the proposed method
retains accurate calibration.  
Furthermore, due to the continuous-time nature of the proposed method compared to the discrete-time baseline,
improved performance is also observed on flat ground. 
\par
The rest of this paper is organized as follows.
Section~\ref{sec:continuous_time} reviews the continuous-time
LiDAR-IMU calibration method used as the base system for the proposed method.
Section~\ref{sec:proposed_method} describes the proposed method
including the novel residuals as well as the measured quantities
that are required. Section~\ref{sec:results} presents results of the proposed approach,
including an ablation and misinitialization studies.
Section~\ref{sec:conclusion} concludes and presents directions for future work.

%% file: sections/proposed_method.tex
\section{Continuous-Time LiDAR-IMU Calibration}
\label{sec:continuous_time}
This section presents a very review of the continuous-time LiDAR-IMU calibration system OA-Calib~\cite{oaLv2022},
used as the base for the proposed method.
The choice of OA-Calib as the base method contrasts with GRIL-Calib, which builds upon the discrete-time system of LI-Init~\cite{zhu2022liinit}.
OA-Calib is a mature continuous-time LiDAR-IMU calibration framework that naturally uses all measurements without downsampling
or finite-differencing, whose improved repeatability on self-collected datasets is shown in~\cite{oaLv2022}.
Since the contribution of the proposed work is in the ground-constraint residuals, the rest of the proposed calibration system
is inherited unchanged from~\cite{oaLv2022}. 
%
%
%
%
%
%
%
\par
OA-Calib utilizes a B-Spline with $\NumCtrlPoints$ control points used to parametrize the IMU trajectory.
Each control point is parametrized by the pose transformation~\cite{barfoot2024state}
\begin{align}
\mbf{T}_{ab_i}=
\begin{bmatrix}
    \mbf{C}_{ab_i} & \mbf{r}_a^{b_ia} \\
    \mbf{0} & 1
\end{bmatrix},    
\end{align}
where $\rframe{b_\CtrlPtIdx}$ denotes the reference frame of the $\CtrlPtIdx$'th control point,
and $b_\CtrlPtIdx$ denotes its physical position. 
The world frame $\rframe{a}$ and the world reference position $a$ are set to be
the IMU reference frame and reference position, respectively, at the time that the first IMU measurement is received.
Formally, the world reference frame
$\rframe{a}=\rframe{b_0}$, and the 
world physical reference point is $a=b_0$. 
%
The control points of the B-Spline are thus given by
$\mc{X}_{\text{spline}}
=\left(
        \mbf{C}_{ab_0}, 
        \mbf{r}_{a}^{b_0 a}, \dots
        \mbf{C}_{ab_\CtrlPtIdx}, 
        \mbf{r}_{a}^{b_\CtrlPtIdx a}, \dots
        \mbf{C}_{ab_\NumCtrlPoints}, 
        \mbf{r}_{a}^{b_\NumCtrlPoints a}; t_i
\right)
$, where the spline time knots $t_i$ are \emph{fixed} throughout the optimization. 
The full optimization state is given by
\begin{align}
    \FullState &= 
    \left(
        \mc{X}_{\text{spline}},
        \mc{X}_{\text{intr}},
        \mbf{b}_{\text{IMU}},
        \mbf{g}_a, 
        \mbf{C}_{bl}, 
        \mbf{r}_{b}^{lb},
        \tau_{bl}
    \right), 
    \label{eq:full_state}
\end{align}
where $\mc{X}_{\text{intr}}$ denotes intrinsic IMU and LiDAR parameters,
$\mbf{b}_{\text{IMU}}$ denotes the IMU accelerometer
and gyroscope biases, assumed constant over the calibration time interval,
$\mbf{g}_a$ is the gravity vector resolved in the world frame,
and $\mbf{C}_{bl},  \mbf{r}_{b}^{lb}, \tau_{bl}$
are the spatiotemporal LiDAR-IMU extrinsics.
The states and constant-bias assumptions are inherited as-is from the OA-Calib system
used as the base for the proposed approach.  
\par
Angular velocity and linear acceleration residuals are obtained by comparing 
IMU measurements to differentiated quantities obtained from the spline. 
Furthermore, a LiDAR residual is constructed using a point-to-plane error. 
The planes
are extracted from the map constructed using the LiDAR.
The points are computed from each LiDAR point measurement that is transformed into the
LiDAR map frame
using the LiDAR-IMU extrinsics and the B-Spline parametrizing the IMU trajectory.
The reader is referred to~\cite{oaLv2022} for details on the continuous-time calibration aspect,
since it is left unchanged in the proposed ground vehicle LiDAR-IMU calibration approach. 
\section{Proposed Method}
\label{sec:proposed_method}
The proposed method consists of extracting ground plane information from the LiDAR point cloud and forming ground plane residuals
that involve calibration parameters. These ground plane residuals are added into the OA-Calib pipeline.
The ground residuals are computed assuming a given time instant.
Therefore, the time subscripts are omitted for clarity in this section.
Furthermore, the pose quantities $\mbf{C}_ab, \mbf{r}_a^{ba}$ at a given time instant
are computed as a function of the spline parameters in the state~\eqref{eq:full_state}.
\par
Three residuals are described and summarized in Table~\ref{tab:residual_summary}.
\begin{table}[t]
\centering
\footnotesize
\caption{Summary of Residuals}
\begin{tabular}{lccc}
\toprule
Residual & Type & Tilted Ground & Uses IMU Inclination \\
\midrule
$\mbf{e}_{\text{flat}}$ & Orientation & No  & No \\
$\mbf{e}_{\text{dist}}$ & Distance    & Yes & No \\
$\mbf{e}_{\text{incl}}$ & Orientation & Yes & Yes \\
\bottomrule
\end{tabular}
\label{tab:residual_summary}
\end{table}
An orientation residual $\mbf{e}_{\text{flat}}$ assuming a flat surface is first proposed,
which assumes colinearity of gravity and surface normal, in the manner of GRIL-Calib. 
Then two residuals are proposed that do not depend on the flat ground assumption. 
A distance residual $\mbf{e}_{\text{dist}}$ is proposed to constrain the extrinsic translation along the
axis of rotation, and an orientation residual $\mbf{e}_{\text{incl}}$ is proposed that uses the ground inclination angle. 
It should be noted that the distance residual $\mbf{e}_{\text{dist}}$ by itself resolves the unobservability of the position extrinsic caused by
the ground vehicle planar motion. 
The inclination residual $\mbf{e}_{\text{incl}}$ provides an additional orientation constraint,
and is included as it is a direct extension to the tilted case of the orientation ground constraint in previous work~\cite{gril2024kim,gctHe2025}.
Ablation studies are presented in Section~\ref{sec:results} that disambiguate the relative contribution of each residual. 
Both $\mbf{e}_{\text{dist}}$ and $\mbf{e}_{\text{incl}}$ are novel residuals that work on tilted ground.
The flat ground orientation residual $\mbf{e}_{\text{flat}}$ carries the same information as
the orientation ground constraint as~\cite{gril2024kim,gctHe2025}, but avoids the creation of extraneous ground frames. 
\par
Each constraint in the following sections may be rewritten as a nonlinear error to be used in a
nonlinear-least-squares solver. The nonlinear errors are written in the form
$\mbf{e}(\mc{X}_{\text{dep}}; \mc{Y}_{\text{meas}})$, 
where $\mc{X}_{\text{dep}}$ denotes the state subset that this error depends on, and 
$\mc{Y}_{\text{meas}}$ denotes the measurements used to construct this error. 
Throughout this section,
$\ura{\tilde{g}} = -\ura{g}/\norm{\ura{g}}$
denotes the unit vector pointing opposite to gravity. 
The normal vector $\ura{n}$ is used to denote the unit normal vector
directed upward from the ground. 
\subsection{Required Quantities and Assumptions}
In the same manner as previous ground-constraint methods~\cite{gril2024kim,gctHe2025},
knowledge of the IMU height $d_b$ is required, corresponding to the shortest distance from the IMU to the local ground plane. 
This quantity is measured physically beforehand,
as it cannot be refined on strictly flat ground.
On strictly planar surfaces, the accuracy of $d_b$ directly affects the accuracy of the position
extrinsic along the axis of rotation, and has to be known as accurately as possible. 
On the LiDAR side, the ground plane is extracted from the LiDAR point cloud
using Patchwork++~\cite{lee2022patchwork}
at each time instant
yielding the surface normal resolved in the LiDAR frame $\mbf{n}_{l}$ as well as
the distance of the LiDAR above the ground $d_{l}$. Therefore, the ground must be reasonably planar
for the ground plane extraction to function. Robustness is provided through
the use of robust loss functions in the calibration optimization to reject outliers. 
\par
Furthermore, the proposed orientation residual $\mbf{e}_{\text{incl}}$ on tilted ground requires knowledge of the ground inclination
$\cos \left( \phi \right) = \ura{\tilde{g}} \cdot \ura{n}$. 
The ground inclination may be obtained from a separate sensor or estimator that provides pitch and roll.
A method for doing so using knowledge of the IMU-resolved gravity on flat ground is provided.  
\par 
\subsection{Orientation Residual Assuming Flat Surface}
\label{sec:orientation_residual_flat_surface}
This subsection describes a flat-ground residual that carries the same information as the orientation residual of~\cite{gril2024kim}.
However, it avoids the introduction of extraneous ground reference frames
that are used in~\cite{gril2024kim}, simplifying practical interpretation and implementation.
\par
If the surface being driven on is flat, the surface normal is colinear with the gravity vector, 
\begin{align}
\ura{n} = \ura{\tilde{g}}, 
\label{eq:surface_normal_colinear_gravity}    
\end{align}
Resolving~\eqref{eq:surface_normal_colinear_gravity} in $\rframe{l}$ yields
\begin{align}
    \mbf{n}_l = \mbf{C}_{bl}^\trans \mbftilde{g}_b, \label{eq:orientation_flat_surface_residual}
\end{align} 
where $\mbftilde{g}_b = \mbf{C}_{ab}\mbftilde{g}_a$ depends on the IMU rotation computed from the B-Spline $\mbf{C}_{ab}$
and the world frame gravity vector $\mbf{g}_a$. 
The corresponding error vector is thus
\begin{align}
    \mbf{e}_{\text{flat}}
    \left(
        \mbf{C}_{ab}, \mbf{g}_a, \mbf{C}_{bl}; \mbf{n}_l  
    \right) &= 
    \mbf{n}_l - \mbf{C}_{bl}^\trans \mbftilde{g}_b \\ 
    &= 
    \mbf{n}_l + \mbf{C}_{bl}^\trans \frac{\mbf{g}_b}{\norm{\mbf{g}_b}_2}.
    \label{eq:flat_ground_residual}
\end{align}
\subsection{Distance Residual} 
The distance residual is constructed by considering the ground plane expressed using each sensor.
It does not require the ground to be flat, and constrains the position extrinsic along the axis of rotation of the vehicle. 
Given the IMU height $d_b$ and the LiDAR height $d_l$,
their difference is equal to the position extrinsic projected onto the
surface normal,  
\begin{align}
     \ura{n} \cdot \ura{r}^{lb} = d_l - d_b, \label{eq:plane_subtracted}
\end{align}
and resolving both sides of ~\eqref{eq:plane_subtracted} in $\rframe{l}$ yields
\begin{align}
    \mbf{r}^{lb^\trans}_l \mbf{n}_l = d_l-d_b, \label{eq:distance_residual_goveq}
\end{align}
which constrains the position extrinsic along the direction of the normal to the surface. 
This constraint is fully independent of the flat ground assumption. 
The choice to resolve in $\rframe{l}$ makes this residual independent of the flat ground assumption, since the surface normal
can be extracted from ground segmentation of the LiDAR point cloud,
in the LiDAR frame regardless of inclination. 
The corresponding error is written 
\begin{align}
     \mbf{e}_{\text{dist}}
     \left(
        \mbf{C}_{bl}, \mbf{r}^{lb}_b; d_b, d_l, \mbf{n}_l 
    \right)
     &= d_l-d_b -
     \left(\mbf{C}_{bl}^\trans \mbf{r}_b^{lb}\right) ^\trans
     \mbf{n}_l.
     \label{eq:distance_residual}
\end{align}
The distance residual constrains the position extrinsic along the surface normal. 
For typical operating conditions, the surface normal corresponds to the axis
of rotation of the vehicle. 
Under this assumption, this directly corresponds to the part of the position extrinsic that is not observable in the single-axis rotation motion
typically experienced by ground vehicles, resolving the position extrinsic unobservability. 
%
%
\subsection{Inclination Residual}
This section proposes an extension of~\eqref{eq:orientation_flat_surface_residual}
to the case where the surface is tilted and the gravity vector is no longer colinear with the surface normal. 
For a given sequence, the inclination residual \emph{or} the flat-ground residual of Section~\ref{sec:orientation_residual_flat_surface}
should be used, depending on whether the ground is assumed flat.
For flat ground, both residuals are valid. However, on tilted ground, the inclination residual remains valid while
the flat-ground residual of Section~\ref{sec:orientation_residual_flat_surface} does not, and they will disagree with each other. 
\par
The proposed inclination residual requires knowledge of the surface inclination, $\cos(\phi)$. 
This quantity has to be provided through a separate sensor or estimation method.
In the present work, it is computed by assumption of knowledge
of the gravity vector resolved in the IMU frame on flat ground, denoted $\mbf{g}_f$, 
together with its normalized version
$\mbftilde{g}_f=-\mbf{g}_f / \norm{\mbf{g}_f}_2$.
The knowledge of $\mbf{g}_f$ is a strong assumption.
In practice, this method of computing the inclination requires
a single initial flat ground calibration sequence.
The resulting $\mbf{g}_f$ can then be reused for
subsequent calibration sequences collected on tilted ground,
under the assumption that the IMU is rigidly mounted on the vehicle and has not shifted. 
In the proposed experiments, it is computed by first running the algorithm on
a known flat-ground sequence, and extracting the optimized $\mbf{g}_a$.
With knowledge of $\mbf{g}_f$, 
the inclination is computed as
\begin{align}
    \cos (\phi)=\mbftilde{g}_b^\trans \mbftilde{g}_f.
    \label{eq:inclination_angle_from_gravities}
\end{align}
%
\par
With $\cos(\phi)$ is provided, the inclination residual can be written as 
\begin{align}
    \cos \left( \phi \right) &= \ura{\tilde{g}} \cdot \ura{n}.
\end{align}
Resolving in $\rframe{l}$ yields
\begin{align}
    \cos \left( \phi \right) 
    &= \mbftilde{g}_b^\trans \mbf{C}_{lb}^\trans \mbf{n}_l.
\end{align}
The corresponding error term is
\begin{align}
  \mbf{e}_{\text{ori}}
    &=
    \cos\left( \phi \right) + \frac{\mbf{g}_a^\trans \mbf{C}_{ab}^\trans
    \mbf{C}_{bl} \mbf{n}_l}{\norm{\mbf{g}_a}_2}.
    \label{eq:error_orientation_tilted}
\end{align}
The inclination residual is written by substituting the inclination angle~\eqref{eq:inclination_angle_from_gravities}
into the orientation residual~\eqref{eq:error_orientation_tilted}, which yields
\begin{align}
    \mbf{e}_{\text{incl}}(
        \mbf{C}_{ab}, \mbf{g}_a, \mbf{C}_{bl}; \mbf{g}_f, \mbf{n}_l
        ) 
    &= 
    \frac{
    \mbf{g}_a^\trans \mbf{C}_{ab}^\trans}{\norm{\mbf{g}_a}_2}
    \left( \mbftilde{g}_f - \mbf{C}_{bl} \mbf{n}_l \right).
    \label{eq:inclination_residual_tilted}
\end{align}
%
\begin{figure}
    \centering
    \includegraphics[width=0.45\columnwidth]{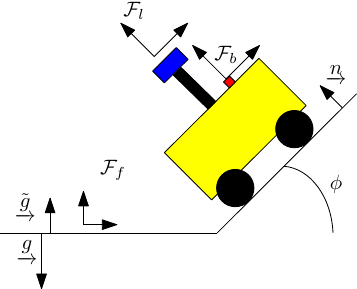}
    \caption{Problem setup: LiDAR-IMU calibration on a tilted planar surface.}
    \label{fig:tilted_plane}
\end{figure}

%% file: sections/results.tex
\section{Results}
\label{sec:results}
The proposed algorithm was compared to GRIL-Calib~\cite{gril2024kim} on a number of datasets.
While the closest work is the continuous-time GCT-Calib~\cite{gctHe2025}, their algorithm
source code is not openly
available for benchmarking, and thus cannot be reasonably reproduced. 
The proposed distance residual requires IMU height, in the same manner as GRIL-Calib.
The inclination residual replaces the flat-ground assumption with the requirement of a surface inclinaiton
measurement. 
The IMU height was measured physically. The IMU-resolved flat-ground gravity vector
was extracted by taking a known flat-ground sequence, running the calibration algorithm withn
the flat-ground assumption, and using the resulting gravity vector. 
Furthermore, both the proposed method and GRIL-Calib require the distance to the ground from the LiDAR,
as well as the normal to the ground resolved in the LiDAR frame.
The Patchwork++ ground segmentation algorithm~\cite{lee2022patchwork} was used to segment the ground and
to extract the distance and surface normal, in the same manner as GRIL-Calib. 
\par
The measured extrinsics, used as an initial guess for the
calibration algorithms, are presented in Table~\ref{tab:all_method_extrinsics_table}.
This is a key difference from the result evaluation in~\cite{gril2024kim}, that used heavily perturbed
measured extrinsics as an initial guess. This is proposed as a more realistic reflection of real-world operation,
where rough sensor placement is known from engineering drawings, and automated procedures are used for refinement. 
\subsection{Metrics}
\label{sec:metrics}
Physical measurements of extrinsic calibration parameters are used for initializing the calibration
procedures, but are of insufficient accuracy to be used as ground truth to evaluate against.
Instead, repeatability is used as a metric for calibration quality. Multiple sequences are collected
from the same sensor rig, and the spread of results across these sequences is
used to assess quality of the calibration algorithm. This spread should nevertheless be interpreted with caution,
as the amount of sequences used is relatively low, ranging between three and seven for each dataset.
Nevertheless, it is proposed that this metric is closest to what would be used by an engineer ``in the wild''
to evaluate the quality of the calibration. 
The measured extrinsic parameters are provided in Table~\ref{tab:all_method_extrinsics_table}.
For the rotation extrinsic, the mean and covariance are computed on the $SO(3)$
group, and converted to Euler angles and degrees for readability in the results tables. 
Formally, given a sequence of results $\left(\mbf{C}_1, \dots, \mbf{C}_\IdxResults, \dots, \mbf{C}_\NumResults \right)$,
the mean rotation in the Riemannian sense is~\cite{moakherMeansAveragingGroup2002}
\begin{align}
    \mbfbar{C} = \argmin_{\mbf{C} \in SO(3)} \sum_{\IdxResults=1}^\NumResults
    \left\| \text{log} \left(\mbf{C}^\trans \mbf{C}_\IdxResults \right) \right\|_\text{F}^2.
    \label{eq:mean_C}
\end{align}
Denoting the differences in the Lie algebra as
$\dxi_\IdxResults = \Log \left(\mbfbar{C}^\trans \mbf{C}_\IdxResults \right)$, it can be shown~\cite{moakherMeansAveragingGroup2002} that 
$\mbfbar{C}$ must satisfy
\begin{align}
    \sum_{\IdxResults=1}^\NumResults \dxi_\IdxResults = 0.
\end{align}
As such, the mean rotation $\mbfbar{C}$ is the rotation matrix that ensures the mean difference in the Lie algebra,
$\dxibar = \frac{1}{\NumResults} \sum_{\IdxResults=1}^\NumResults \dxi_\IdxResults$, is zero.
An algorithm for computing $\mbfbar{C}$ is provided in~\cite{manton2004globally}.
The empirical covariance matrix is computed as
\begin{align}
    \mbs{\Sigma} &= 
    \frac{1}{\NumResults-1} \sum_{\IdxResults=1}^\NumResults
    \dxi_\IdxResults
    \dxi_\IdxResults^\trans,
\end{align} 
with the per-component standard deviation given by
\begin{align}
    \mbs{\sigma} = 
    \begin{bmatrix}
        \sqrt{\Sigma_{11}} & 
        \sqrt{\Sigma_{22}} & 
        \sqrt{\Sigma_{33}}  
    \end{bmatrix}.
    \label{eq:per_component_std}
\end{align}
In subsequent tables, the mean rotation entry corresponds to the $1-2-3$ Euler angle decomposition
of $\mbfbar{C}$ in~\eqref{eq:mean_C},
while the standard deviations used to quantify result spread correspond to $\mbs{\sigma}$ in~\eqref{eq:per_component_std}.
Both the mean rotation Euler angles and the standard deviations are normalized to units of degrees. 
A lower standard deviation demonstrates improved repeatability across
calibration sequences, which is taken to be the
performance metric of the tested calibration methods. 
While the standard deviations are considered the metric for performance,
the mean calibration is also reported as a sanity check to compare to
the previously measured extrinsics for each dataset. 
\par
The mean calibration results for each dataset and proposed method are
consolidated in Table~\ref{tab:all_method_extrinsics_table} and discussed below.
\begin{table}
\centering
\caption{Estimated Extrinsics Across All Datasets}
\scriptsize
\label{tab:all_method_extrinsics_table}
\begin{tabular}{|l|c|c|}
\hline
    Method & $\mbf{r}_b^{lb}$ (cm) & $\mbf{C}_{bl}$ (Euler, deg) \\
\hline
\multicolumn{3}{|c|}{Husky UGV, Flat Ground} \\
\hline
    Measured & [24.77, -6.99, -46.36] & [180.00, 0.00, -90.00] \\
    GRIL-Calib     & [25.17, -3.38, -43.55] & [179.98, -0.21, -90.11] \\
    Proposed & [26.00, -7.20, -43.53] & [179.98, -0.17, -87.48] \\
\hline
\multicolumn{3}{|c|}{Husky UGV, Incline (Note: GRIL-Calib Diverges.)} \\
\hline
    Measured & [24.77, -6.99, -46.36] & [180.00, 0.00, -90.00] \\
    GRIL-Calib     & [31.29, 17.95, -46.43] & [182.17, -1.54, 199.22] \\
    Proposed & [25.76, -8.26, -43.19] & [179.93, -0.15, -87.07] \\
\hline
\multicolumn{3}{|c|}{M2DGR} \\
\hline
    Measured & [27.26, 0.00, 17.95] & [0.00, 0.00, 0.00] \\
    GRIL-Calib     & [25.61, 6.71, 17.81] & [-0.37, -0.08, 0.54] \\
    Proposed & [25.00, 0.52, 17.36] & [-0.47, 0.98, -0.77] \\
\hline
\multicolumn{3}{|c|}{Offroad} \\
\hline
    Measured & [-63.40, 132.60, 34.05] & [0.06, 0.00, -0.26] \\
    GRIL-Calib     & [-61.14, 133.61, 34.83] & [0.15, 0.43, -0.68] \\
    Proposed & [-59.82, 132.72, 33.78] & [0.18, 0.70, -1.25] \\
\hline
\end{tabular}
\end{table}
\subsection{Husky Clearpath Ground Vehicle}
\begin{figure}
    \centering
    \includegraphics[width=0.5\columnwidth]{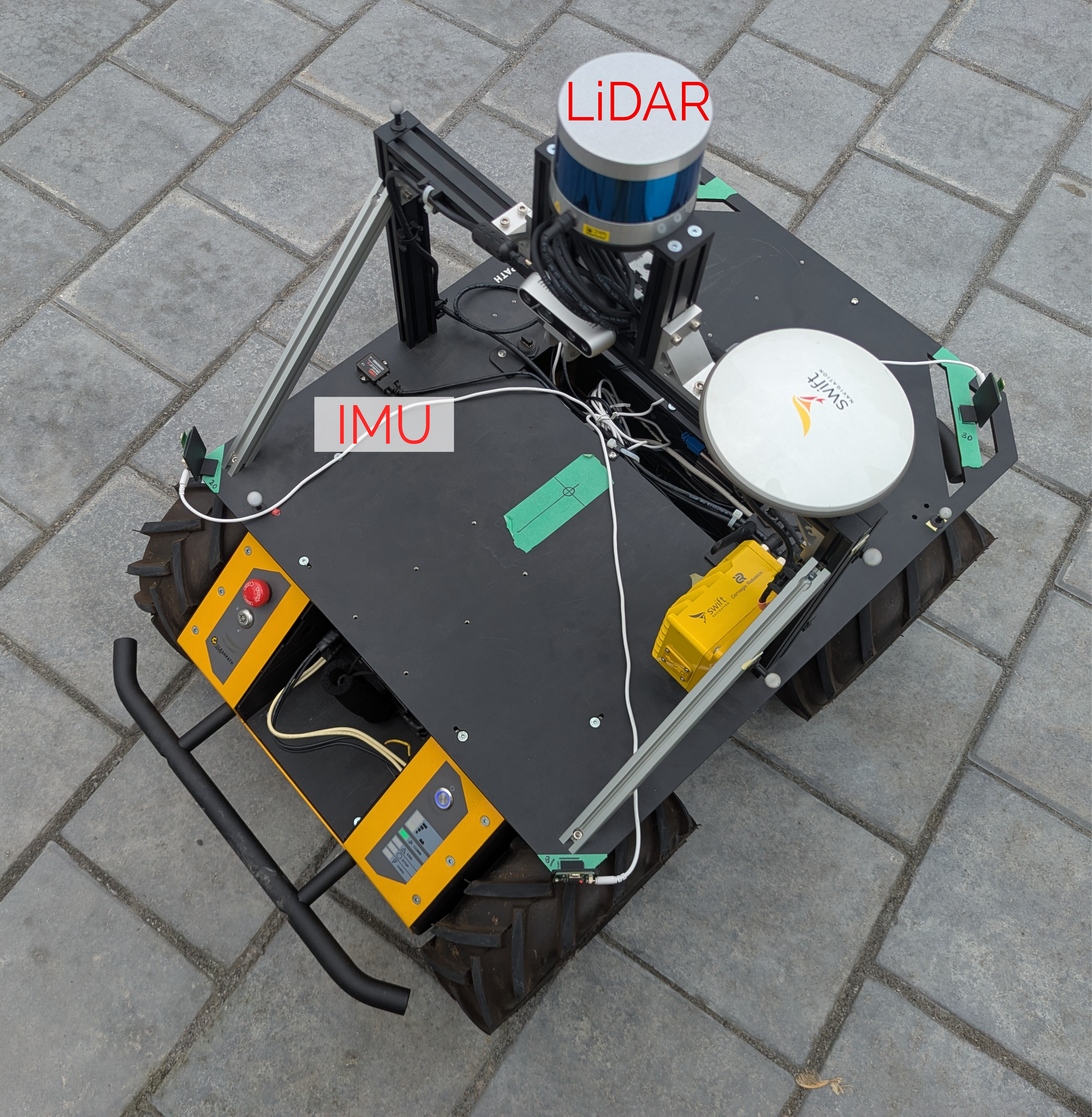}
    \caption{Husky UGV used for algorithm validation.}
    \label{fig:husky_image}
\end{figure}

A Clearpath Husky unmanned ground vehicle equipped with a Velodyne VLP-16 LiDAR and
a 3DM-GX5-AHRS attitude-heading-reference-system for IMU data, pictured in Figure~\ref{fig:husky_image},
was driven on flat and inclined ground
to validate the proposed algorithm. 
Three sequences are collected indoors on flat ground in a lab environment to demonstrate
nominal flat-ground performance for both methods. 
Seven other sequences are collected on inclined ground in an outdoors urban environment on a university campus to demonstrate
performance improvement on inclined ground.  
The previously measured extrinsics are obtained from a visual inspection for the rotation extrinsic, 
and a tape measure for the position extrinsic.
\par
\begin{figure}
    \centering
    \includegraphics[width=0.8\columnwidth]{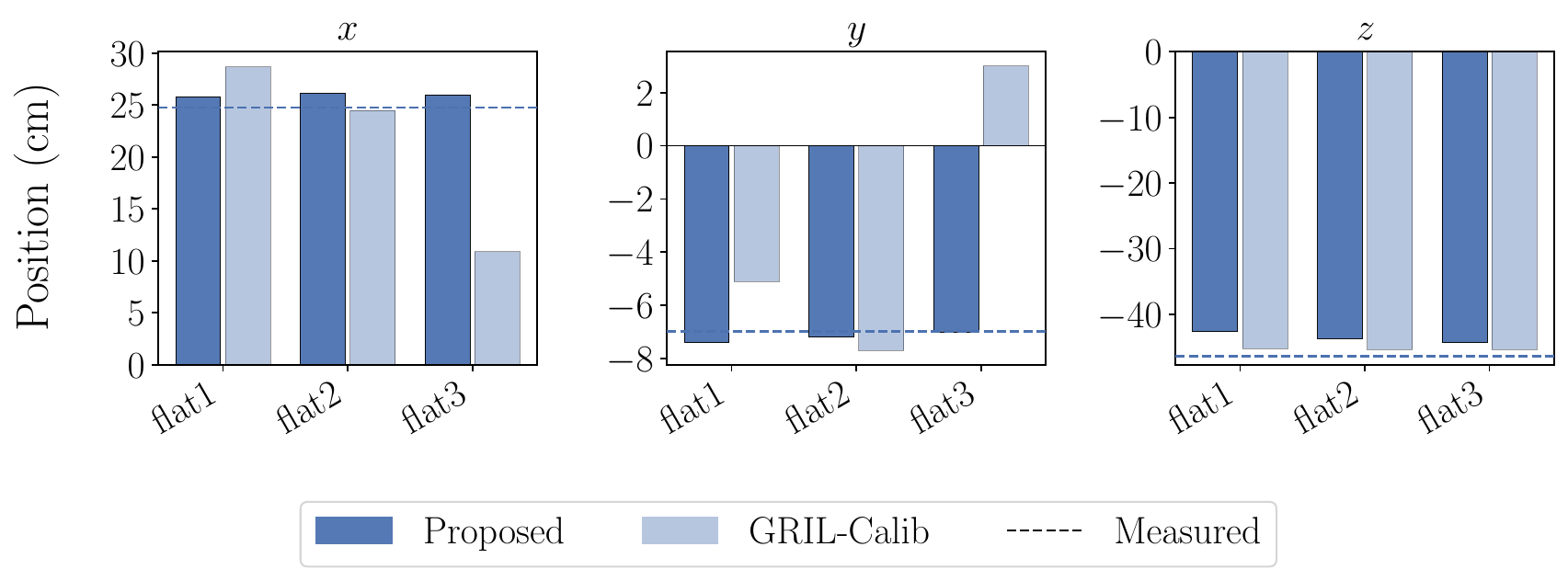}
    \caption{Per-sequence position extrinsic results for the flat-ground Husky UGV dataset.}
    \label{fig:husky_flat_position_extrinsic_per_sequence}
\end{figure}
\begin{figure}
    \centering
    \includegraphics[width=0.7\columnwidth]{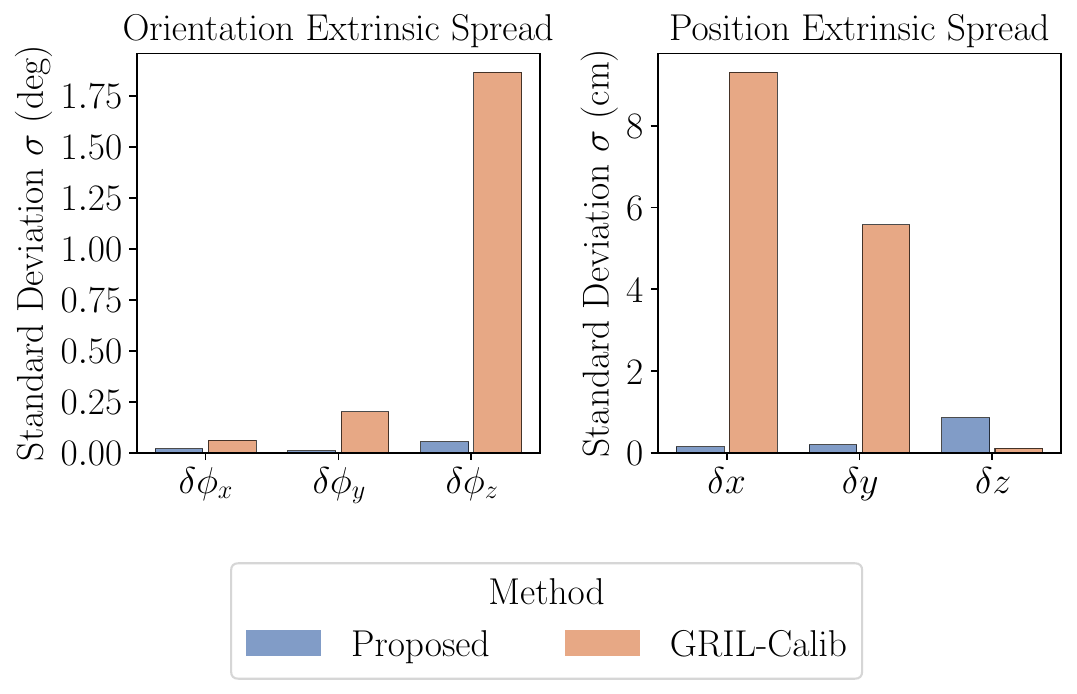}
    \caption{Extrinsic spread for the flat-ground Husky UGV dataset.}
    \label{fig:husky_flat_spread}
\end{figure}

The calibration results for the flat-ground sequences are presented in Figure~\ref{fig:husky_flat_position_extrinsic_per_sequence}.
The extrinsic spread is plotted in Figure~\ref{fig:husky_flat_spread}.
The computed mean extrinsics are presented in Table~\ref{tab:all_method_extrinsics_table}.
Both GRIL-Calib and the proposed method converge to values reasonably close to the
measured extrinsics from Table~\ref{tab:all_method_extrinsics_table}. 
The proposed method achieves increased repeatability across all extrinsic parameters.
Since the sequences are collected on flat ground,
the performance improvement is attributed to the continuous-time
nature of the overall system. 
The performance increase is especially pronounced in the yaw orientation and the $x-y$ position
extrinsics. This is likely due to the planar motion excitation in the vehicle.
The sensors are mounted flat on the vehicle, with yaw and $x-y$ directions
corresponding to the vehicle motion directions.
\par
The results for the inclined ground sequences are presented in
Table~\ref{tab:husky_outdoors_by_sequence}, and are the primary evidence
for the performance improvement of the proposed method on inclined ground.
\begin{table}
\centering
\scriptsize
\begin{tabular}{|l|c|c|}
\hline
\multicolumn{3}{|c|}{Proposed} \\ 
\hline
& $\mbf{r}_b^{lb}$ (cm) & $\mbf{C}_{bl}$ (Euler, deg)  \\
\hline
    Outdoors1 & [25.600, -7.700, -46.600] & [179.965, -0.179, -88.440] \\
    Outdoors2 & [25.400, -8.200, -41.000] & [179.969, -0.029, -87.740] \\
    Outdoors3 & [29.300, -5.400, -46.700] & [179.638, -0.306, -87.790] \\
    Outdoors4 & [23.700, -14.100, -42.400] & [-179.983, -0.103, -82.450] \\
    Outdoors5 & [26.600, -8.100, -43.600] & [180.000, -0.150, -86.010] \\
    Outdoors6 & [24.500, -8.300, -41.800] & [179.947, -0.179, -89.120] \\
    Outdoors7 & [24.700, -6.100, -40.200] & [-179.996, -0.160, -87.700] \\
\hline
\multicolumn{3}{|c|}{GRIL-Calib} \\ 
\hline
& $\mbf{r}_b^{lb}$ (cm) & $\mbf{C}_{bl}$ (Euler, deg)  \\
\hline
    Outdoors1 & [49.304, -9.349, -46.908] & [-179.746, 3.644, 77.519] \\
    Outdoors2 & [-6.948, 16.054, -42.245] & [179.561, -3.068, -69.199] \\
    Outdoors3 & [29.370, 51.323, -51.922] & [-174.973, -0.909, 155.730] \\
    Outdoors4 & [-13.562, 50.277, -47.859] & [-175.170, 0.754, 136.414] \\
    Outdoors5 & [99.302, 8.624, -47.950] & [177.023, -1.507, -173.926] \\
    Outdoors6 & [34.369, 2.119, -43.992] & [-179.190, -0.707, -89.989] \\
    Outdoors7 & [27.174, 6.580, -44.160] & [178.368, -1.326, -81.869] \\
\hline
\end{tabular}
\caption{Results for each sequence of inclined Husky UGV dataset}
\label{tab:husky_outdoors_by_sequence}
\end{table}
Direct comparison to the measured extrinsic of Table~\ref{tab:all_method_extrinsics_table} shows that the GRIL-Calib method
yields an incorrect answer, while the proposed method still converges to values that are reasonably close both to
the measured extrinsics of Table~\ref{tab:all_method_extrinsics_table}
and the mean flat-ground calibration results of Figure~\ref{tab:all_method_extrinsics_table}. 
\subsection{M2DGR}
The M2DGR dataset~\cite{yin2022m2dgr}, equipped with a Velodyne VLP-32C and Handsfree A9 IMU, collected on flat ground, is used to demonstrate the
advantages of the continuous-time nature of the proposed method.
The sequences used
and their respective time segments
are chosen to directly correspond to the ones used in the analysis of GRIL-Calib~\cite{gril2024kim}.
The previously measured extrinsics are extracted from the calibration provided by
the dataset paper~\cite{yin2022m2dgr}. 
\par
\begin{figure}
    \centering
\includegraphics[width=\columnwidth]{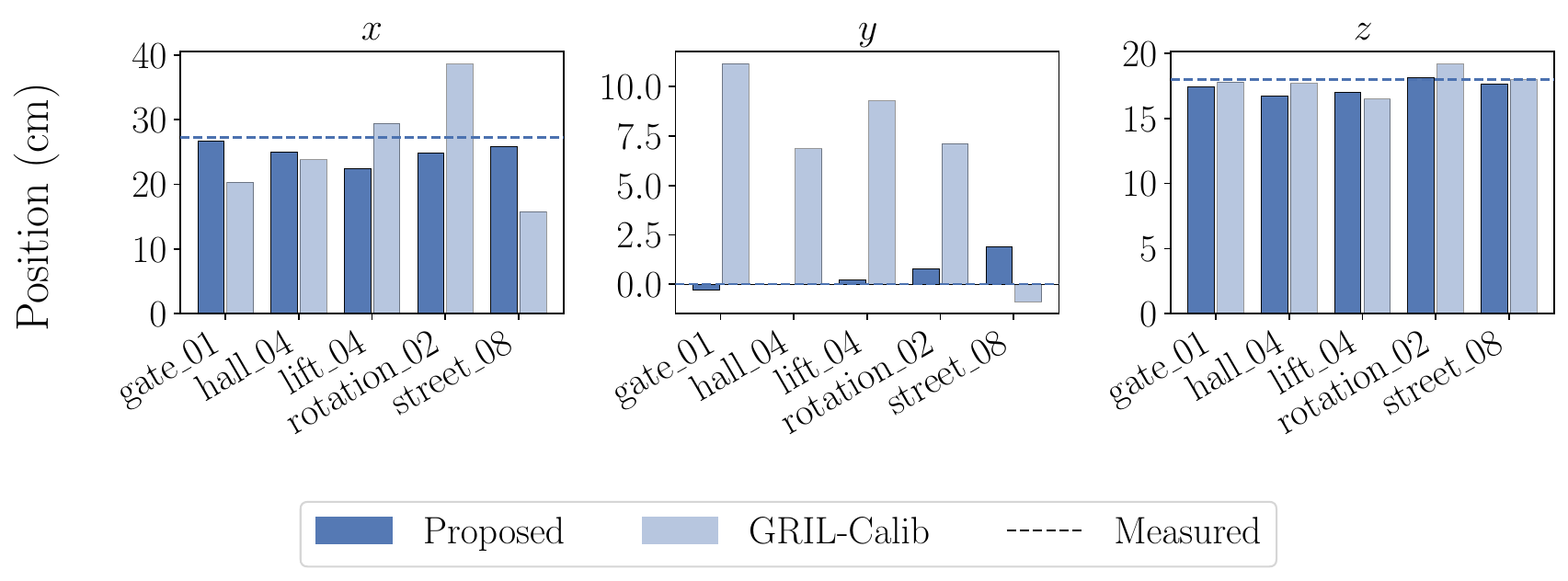}
\caption{Per-sequence position extrinsic results for the M2DGR dataset.}
\label{fig:m2dgr_position_extrinsic_per_sequence}
\end{figure}
\begin{figure}
\centering
\includegraphics[width=0.7\columnwidth]{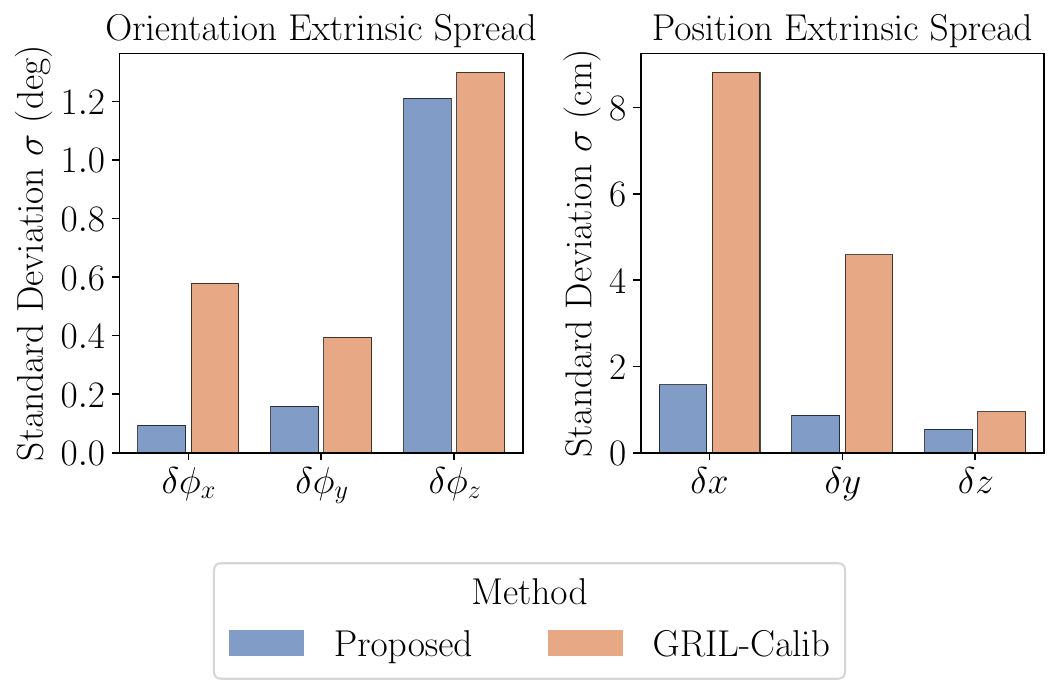}
\caption{Extrinsic spread for the M2DGR dataset.}
\label{fig:m2dgr_spread}
\end{figure}

The per-sequence position extrinsics, the spread in the extrinsics, and the mean computed extrinsics
are presented in Figure~\ref{fig:m2dgr_position_extrinsic_per_sequence},
Figure~\ref{fig:m2dgr_spread}, and Table~\ref{tab:all_method_extrinsics_table}, respectively. 
\par 
The mean extrinsics of both methods in Table~\ref{tab:all_method_extrinsics_table}
are comparable to the measured extrinsics of Table~\ref{tab:all_method_extrinsics_table}. 
The $x-y$ position extrinsic
repeatability for the proposed is significantly improved compared to GRIL-Calib in Figure~\ref{fig:m2dgr_spread}, mirroring
the results obtained on the Husky UGV flat-ground sequences.
The spread change is mostly driven by the \texttt{street\_08}
and \texttt{rotation\_02} sequences, for the $x$ and $y$ position extrinsic respectively
as seen from Figure~\ref{fig:m2dgr_position_extrinsic_per_sequence}.  
The yaw extrinsic repeatability is comparable; however it is slightly degraded compared to the GRIL-Calib baseline.
Both methods yield about one degree of standard deviation in the yaw calibration extrinsic. 
This differs from the Husky UGV sequences, where the proposed method had a standard deviation of a tenth of a degree.
It is also worth noting that the GRIL-Calib spread in Figure~\ref{fig:m2dgr_spread}
is comparable in magnitude
compared to the RMSE values presented in~\cite{gril2024kim}. 
%
\subsection{Offroad Vehicle Dataset}
\begin{figure}
    \centering
    \includegraphics[width=\columnwidth]{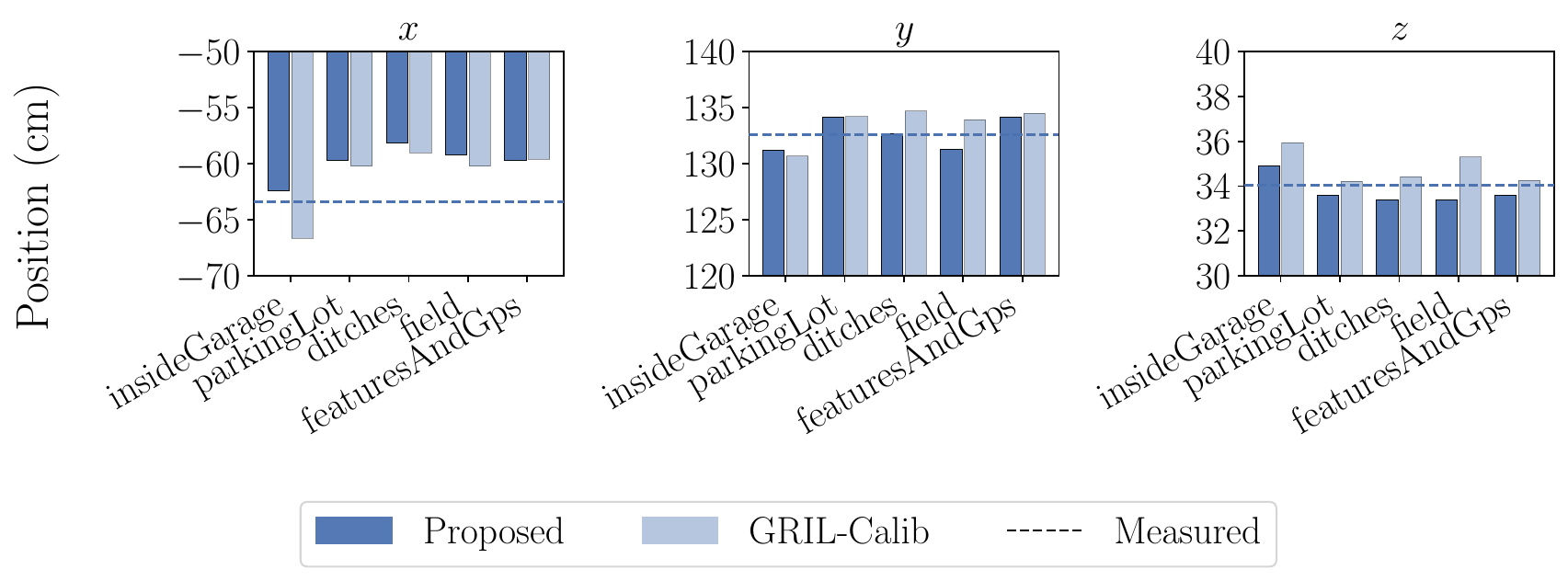}
    \caption{Per-sequence position extrinsic results for the Offroad dataset.}
    \label{fig:offroad_position_extrinsic_per_sequence}
    \end{figure}

\begin{figure}
\centering
\includegraphics[width=0.7\columnwidth]{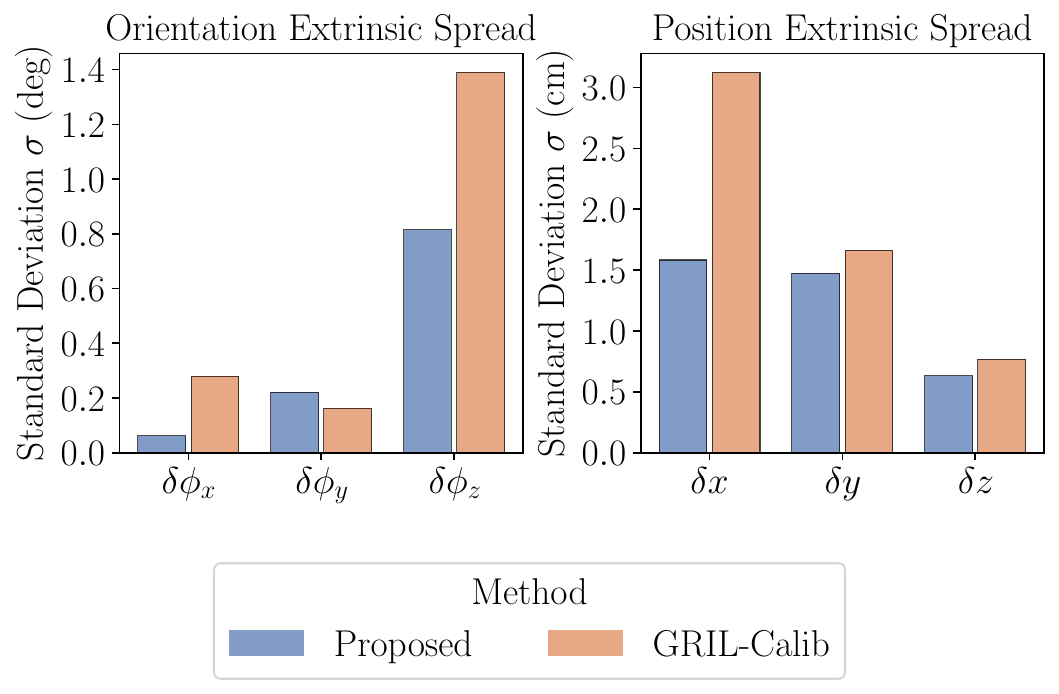}
\caption{Extrinsic spread for the Offroad dataset.}
\label{fig:offroad_spread}
\end{figure}

\begin{figure}
\centering
\includegraphics[width=0.55\columnwidth]{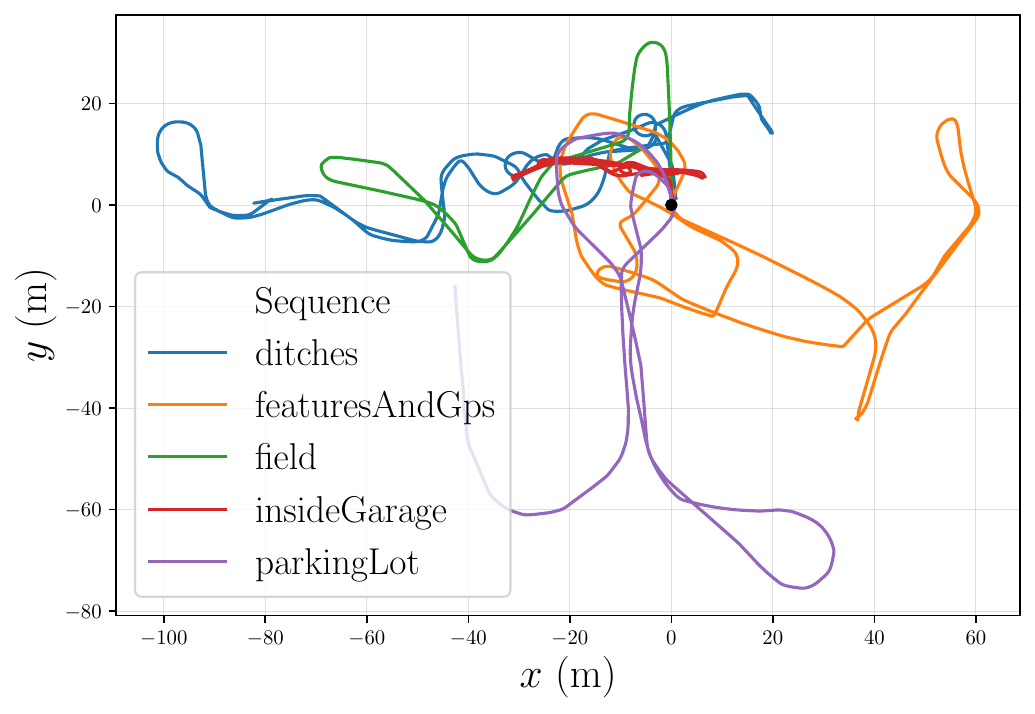}
\caption{Offroad dataset trajectory visualization.}
\label{fig:offroad_trajectory_visualization}
\end{figure}

An offroad vehicle equipped with an Ouster OS1-32 LiDAR and a XSENS MTI-200-2A8G4 IMU
was driven in various environments, both indoors and outdoors.
The mean extrinsics of the two methods, presented in Table~\ref{tab:all_method_extrinsics_table} are very close to each other.
The spread in the obtained extrinsics is improved by the proposed method across the board,
as shown in Figure~\ref{fig:offroad_spread}, except for the $y$-component of the orientation error that is slightly worsened. 
Per-sequence position extrinsic results are presented in Figure~\ref{fig:offroad_position_extrinsic_per_sequence}, where the dashed line
shows the measured extrinsic used to initialize the calibration.
The \texttt{insideGarage} sequence shows deviation with respect to the other sequences. 
It is also worth noting that the $z$-component of the position extrinsic in Figure~\ref{fig:offroad_position_extrinsic_per_sequence}
is systematically higher for GRIL-Calib than for the proposed method. Both methods use the same Patchwork++ code~\cite{lee2022patchwork}
with the same configuration parameters, which means that the bias is due to some other convergence properties
of the method.  
A top-down visualization of the offroad trajectories is provided in Figure~\ref{fig:offroad_trajectory_visualization}, where the trajectory
starting position is normalized to zero for all sequences.  

\subsection{Ablation Study on Offroad Dataset}
\label{sec:ablation_offroad}
\begin{figure}
\centering
\includegraphics[width=0.8\columnwidth]{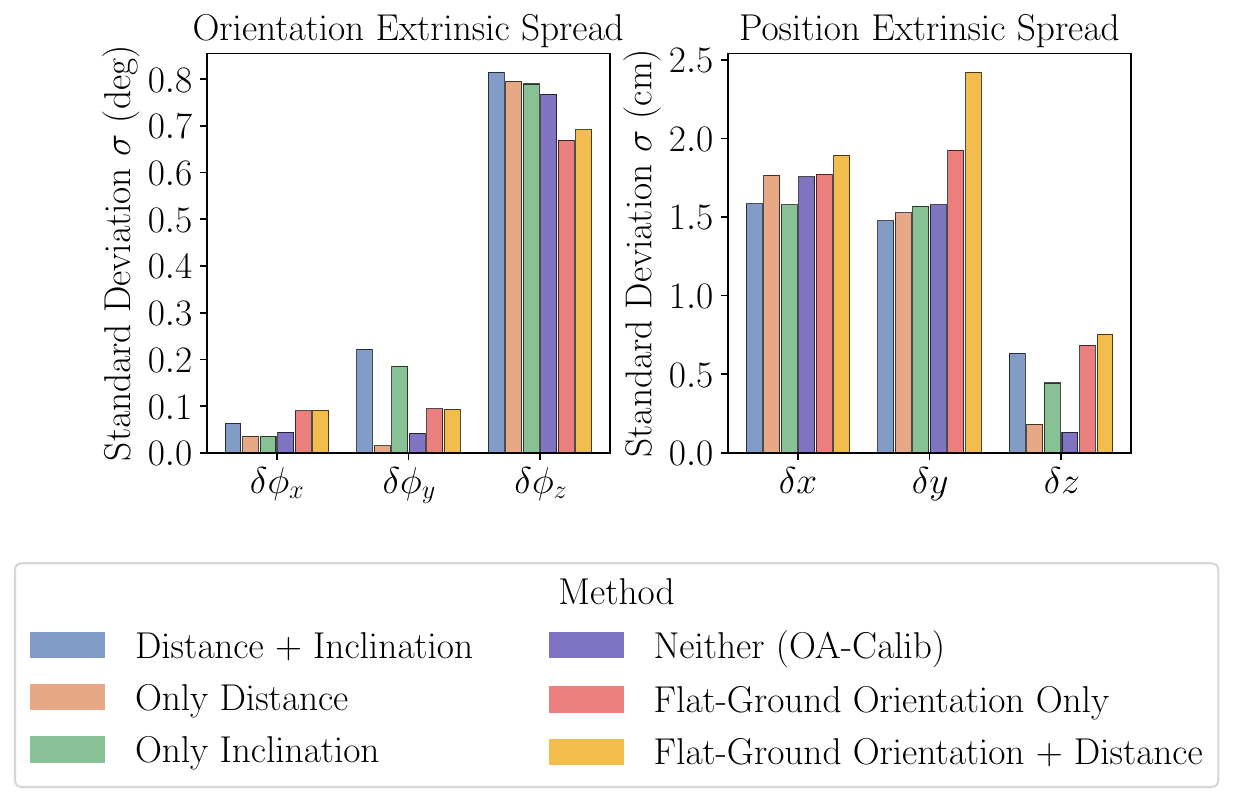}
\caption{Extrinsic spread for the Offroad (ablation) dataset.}
\label{fig:ablation_offroad_spread}
\end{figure}

\begin{figure}
    \centering
\includegraphics[width=\columnwidth]{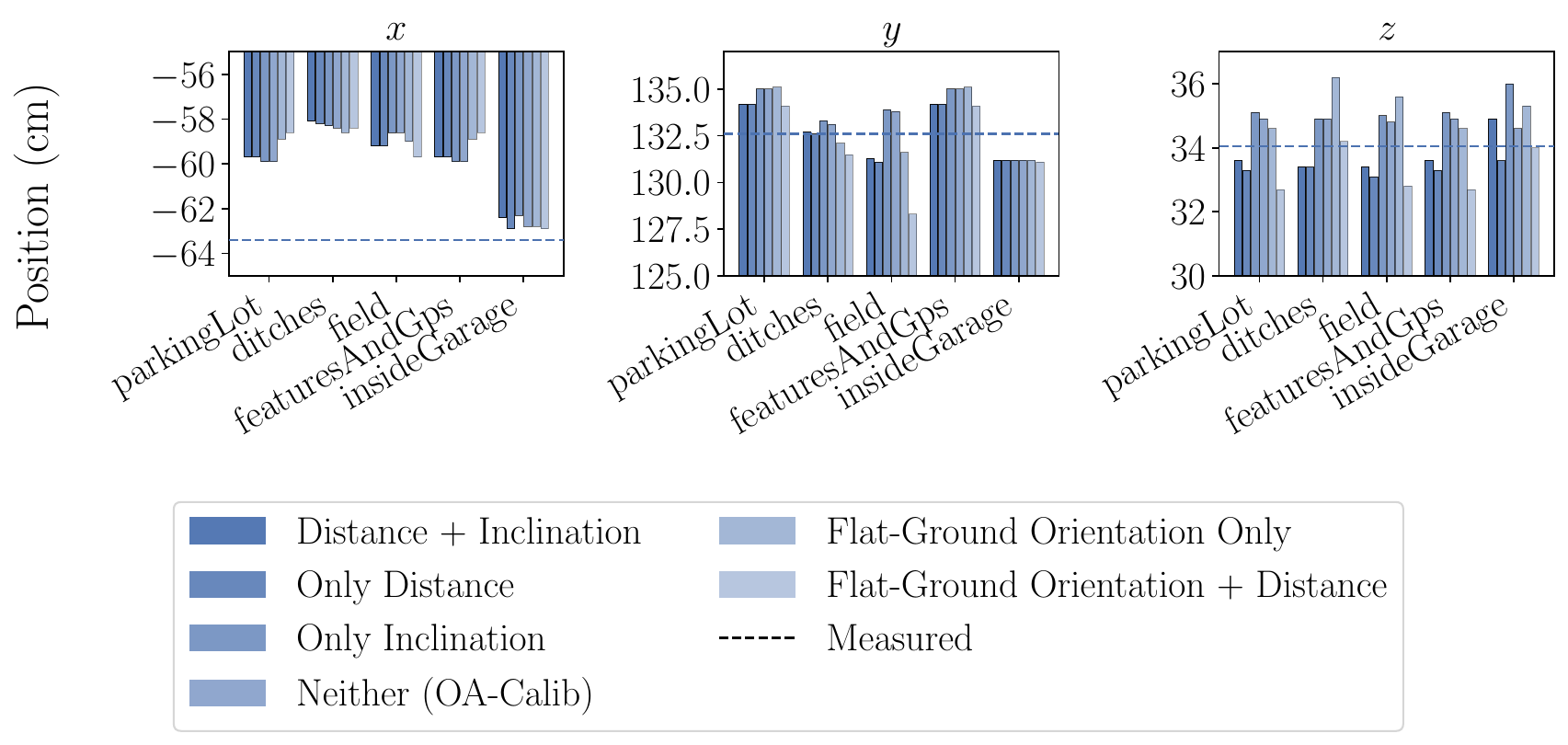}
\caption{Per-sequence position extrinsic results for the Offroad (ablation) dataset.}
\label{fig:ablation_offroad_position_extrinsic_per_sequence}
\end{figure}

An ablation study was run on the offroad dataset
in order to
analyze the relative contribution of each residual to the calibration solution.
The results are presented in
Figure~\ref{fig:ablation_offroad_spread} 
and Figure~\ref{fig:ablation_offroad_position_extrinsic_per_sequence}
for the extrinsic parameter spread and per-sequence position extrinsic values, respectively.
The compared methods correspond to combinations of the flat-ground~\eqref{eq:flat_ground_residual}
and inclined orientation residuals~\eqref{eq:error_orientation_tilted}, and
the distance residual~\eqref{eq:distance_residual}. 
The flat-ground residuals yield significantly higher spread in the position extrinsic,
which seems to be significantly driven by the \texttt{fields} sequence
of Figure~\ref{fig:ablation_offroad_position_extrinsic_per_sequence}.
The baseline of OA-Calib has the lowest spread among the proposed methods.
This is conjectured to be caused by the initialization of the method, where the method is initialized
with the pre-existing measured extrinsics, and does not drift much from that measured extrinsic. 
This also shows the limitations of the repeatability metric used in this work. OA-Calib cannot by itself
refine the axis-of-rotation position extrinsic, as it is not observable without ground constraints.
However, the spread in Figure~\ref{fig:ablation_offroad_spread} shows that any spurious updates present in
the nonlinear-least-squares solver are quite low in magnitude.
The $z$-extrinsic remains essentially unchanged from the original value of $34.05\ \text{cm}$ for all sequences.   
%
\subsection{Offroad With Misinitialized Extrinsics}
\begin{figure}
\centering
\includegraphics[width=0.7\columnwidth]{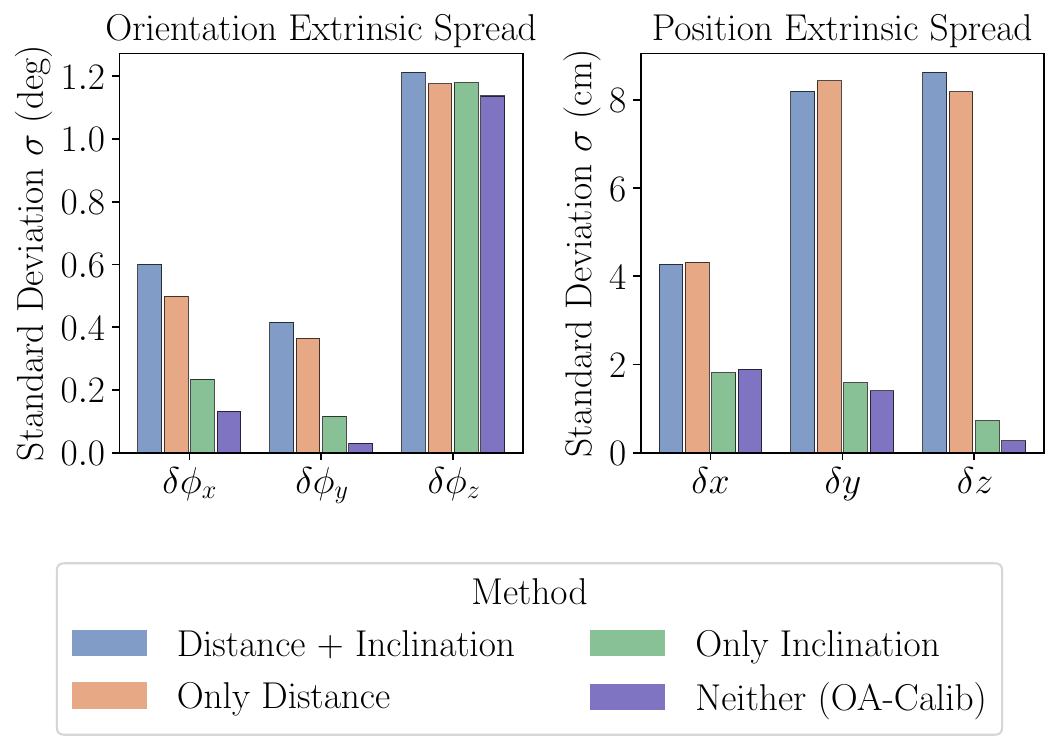}
\caption{Extrinsic spread for ablation study on offroad dataset with misinitalized extrinsics.}
\label{fig:ablation_offroad_misinitialized_spread}
\end{figure}

\begin{figure}
\centering
\includegraphics[width=\columnwidth]{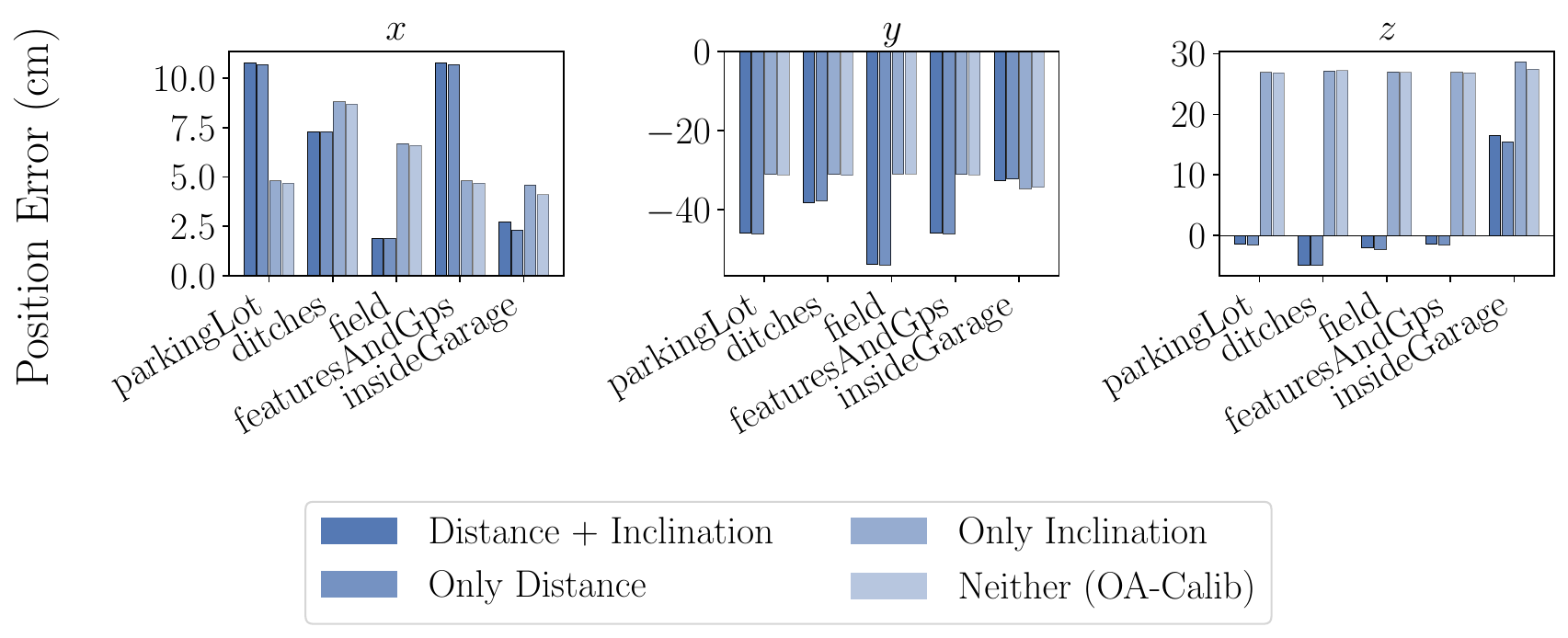}
\caption{Per-sequence position error for ablation study on offroad dataset with misinitalized extrinsics.}
\label{fig:position_error_offroad_misinitialized_extrinsics}
\end{figure}

The results of Section~\ref{sec:ablation_offroad} show that the system without any additional constraints performs
better in terms of repeatability, which shows the limits of repeatability as a metric.
The updates to an unobservable state direction in calibration are due to spurious updates from the solver,
and Section~\ref{sec:ablation_offroad} showed that the $z$-position extrinsic, 
which correponds to an unobservable direction for this calibration problem,
remains relatively close to its initial value. 
This section shows that for misinitialized extrinsics, the calibration algorithm without a ground constraint
does not update the unobservable direction. 
To study the effect of misinitialization, the offroad dataset was run with
a misinitialized position extrinsic, set to
$
\mbf{r}_b^{lb}
=
\begin{bmatrix}
-0.6&  1.0&  0.6
\end{bmatrix}
$ meters.
Figure~\ref{fig:ablation_offroad_misinitialized_spread} shows the spread in the resultant
calibration extrinsics. The orientation extrinsic remains unchanged. However, the position extrinsic
demonstrates a far larger spread for cases where the distance residual is used, reflecting the
convergence behaviour of an algorithm initialized far from the true value.
While the $z$-position extrinsic has the smallest spread for the ``Neither''
and ``Only Inclination''
cases, where the distance residual is not used, 
Figure~\ref{fig:position_error_offroad_misinitialized_extrinsics}
shows that this value is heavily biased.
Figure~\ref{fig:position_error_offroad_misinitialized_extrinsics} plots the
position errors with respect to the measured values in Table~\ref{tab:all_method_extrinsics_table}.
The error of around $25$ centimeters in the $z$-extrinsic correponds to the
$z$-extrinsic not being refined from its initial value in the case of
residuals that do not use the distance residual.
Furthermore, the \texttt{insideGarage} sequence shows a large error even for the case of the distance residual.
This is caused by the system setup. The distance residual is wrapped inside of a robust Cauchy loss~\cite{barfoot2024state}.
Depending on how far the system is initialized and what shape parameter is used for the robust loss,
the robust loss may cause entire rejection of the distance residual update.
The exact parameter choice depends on the problem setup. A lenient robust loss may cause updates
from spurious ground measurements, while a harsh robust loss may cause the system to favour misinitialized values. 
%

%% file: sections/conclusion.tex
\section{Conclusion}
\label{sec:conclusion}
This paper presents a 
continuous-time LiDAR-IMU targetless calibration method for planar motion on non-flat ground.
Compared to the state-of-the-art flat ground calibration system, the proposed method demonstrates
robustness to tilted ground. Improved performance is also demonstrated on flat ground due to
the continuous-time nature of the proposed method. 
However, the method has limitations.  
The IMU height still has to be known for the distance residual, and the IMU-resolved gravity vector on
flat ground has to be known to compute the inclination of the vehicle. 
Future work includes treating the IMU height as a state parameter instead of a measured, fixed quantity
as well as different methods to compute inclination. 
Furthermore, the lack of precise ground truth, the
difficulty of measuring impact of calibration on estimation accuracy, as well
as sensitivity of the calibration algorithm to initialization makes evaluating the quality of
calibration algorithms, and the resulting calibration, a non-trivial task.
While this work advocates for repeatability as a primary metric, the misinitialization study also shows that
previously measured values are necessary as a sanity check. This shows that robust metrics of calibration
quality, whether based on error, repeatability, or downstream estimator performance,
are an important direction of future research. 
%

%% file: sections/acknowledgements.tex
\section{Acknowledgements}
The authors would like to acknowledge Nicholas Dahdah for help with the dataset collection
and Mitchell Cohen for valuable discussions.

%% file: library.bib
@article{gril2024kim,
  author   = {Kim, TaeYoung and Pak, Gyuhyeon and Kim, Euntai},
  journal  = {IEEE Robotics and Automation Letters},
  title    = {GRIL-Calib: Targetless Ground Robot IMU-LiDAR Extrinsic Calibration Method Using Ground Plane Motion Constraints},
  year     = {2024},
  volume   = {9},
  number   = {6},
  pages    = {5409-5416},
  doi      = {10.1109/LRA.2024.3392081}
}

@article{slope2025xiao,
  author   = {Xiao, Jian and Guo, Jiming and Ouyang, Chenhao and Shi, Junbo and Xu, Yi and Wu, Mengqi},
  journal  = {IEEE Transactions on Instrumentation and Measurement},
  title    = {A Slope-Based Targetless Extrinsic Calibration Method for LiDAR-IMU Systems on Ground Vehicles},
  year     = {2025},
  volume   = {74},
  number   = {},
  pages    = {1-14},
  doi      = {10.1109/TIM.2025.3580830}
}

@article{moakherMeansAveragingGroup2002,
  author  = {Moakher, Maher},
  title   = {Means and Averaging in the Group of Rotations},
  journal = {SIAM Journal on Matrix Analysis and Applications},
  volume  = {24},
  number  = {1},
  pages   = {1-16},
  year    = {2002},
  doi     = {10.1137/S0895479801383877}
}

@inproceedings{manton2004globally,
  author    = {Manton, J.H.},
  booktitle = {ICARCV 2004 8th Control, Automation, Robotics and Vision Conference, 2004.},
  title     = {A globally convergent numerical algorithm for computing the centre of mass on compact Lie groups},
  year      = {2004},
  volume    = {3},
  number    = {},
  pages     = {2211-2216 Vol. 3},
  doi       = {10.1109/ICARCV.2004.1469774}
}

@inproceedings{gctHe2025,
  author    = {He, Tongsheng and Wang, Ping and Das, Debasis and Maharaj, B. T.},
  booktitle = {2025 5th International Conference on Communication Technology and Information Technology (ICCTIT)},
  title     = {GCT-LICalib: Targetless Online LiDAR–IMU Calibration via Ground Plane Motion Constraints and Continuous-Time Optimization},
  year      = {2025},
  volume    = {},
  number    = {},
  pages     = {238-243},
  doi       = {10.1109/ICCTIT68197.2025.11406395}
}

@article{oaLv2022,
  author   = {Lv, Jiajun and Zuo, Xingxing and Hu, Kewei and Xu, Jinhong and Huang, Guoquan and Liu, Yong},
  journal  = {IEEE Transactions on Robotics},
  title    = {Observability-Aware Intrinsic and Extrinsic Calibration of LiDAR-IMU Systems},
  year     = {2022},
  volume   = {38},
  number   = {6},
  pages    = {3734-3753},
  doi      = {10.1109/TRO.2022.3174476}
}

@inproceedings{zhu2022liinit,
  author    = {Zhu, Fangcheng and Ren, Yunfan and Zhang, Fu},
  booktitle = {2022 IEEE/RSJ International Conference on Intelligent Robots and Systems (IROS)},
  title     = {Robust Real-time LiDAR-inertial Initialization},
  year      = {2022},
  volume    = {},
  number    = {},
  pages     = {3948-3955},
  doi       = {10.1109/IROS47612.2022.9982225}
}

@inproceedings{lee2022patchwork,
  author    = {Lee, Seungjae and Lim, Hyungtae and Myung, Hyun},
  booktitle = {2022 IEEE/RSJ International Conference on Intelligent Robots and Systems (IROS)},
  title     = {Patchwork++: Fast and Robust Ground Segmentation Solving Partial Under-Segmentation Using 3D Point Cloud},
  year      = {2022},
  volume    = {},
  number    = {},
  pages     = {13276-13283},
  doi       = {10.1109/IROS47612.2022.9981561}
}

@article{chen2025ikalibr,
  author   = {Chen, Shuolong and Li, Xingxing and Li, Shengyu and Zhou, Yuxuan and Yang, Xiaoteng},
  journal  = {IEEE Transactions on Robotics},
  title    = {iKalibr: Unified Targetless Spatiotemporal Calibration for Resilient Integrated Inertial Systems},
  year     = {2025},
  volume   = {41},
  number   = {},
  pages    = {1618-1638},
  doi      = {10.1109/TRO.2025.3532506}
}

@article{yang2023online,
  author   = {Yang, Yulin and Geneva, Patrick and Zuo, Xingxing and Huang, Guoquan},
  journal  = {IEEE Transactions on Robotics},
  title    = {Online Self-Calibration for Visual-Inertial Navigation: Models, Analysis, and Degeneracy},
  year     = {2023},
  volume   = {39},
  number   = {5},
  pages    = {3479-3498},
  doi      = {10.1109/TRO.2023.3275878}
}

@inproceedings{lee2024degenerate,
  author    = {Lee, Woosik and Chen, Chuchu and Huang, Guoquan},
  booktitle = {2024 IEEE International Conference on Robotics and Automation (ICRA)},
  title     = {Degenerate Motions of Multisensor Fusion-based Navigation},
  year      = {2024},
  volume    = {},
  number    = {},
  pages     = {8113-8119},
  doi       = {10.1109/ICRA57147.2024.10610255}
}

@article{yin2022m2dgr,
  author   = {Yin, Jie and Li, Ang and Li, Tao and Yu, Wenxian and Zou, Danping},
  journal  = {IEEE Robotics and Automation Letters},
  title    = {M2DGR: A Multi-Sensor and Multi-Scenario SLAM Dataset for Ground Robots},
  year     = {2022},
  volume   = {7},
  number   = {2},
  pages    = {2266-2273},
  doi      = {10.1109/LRA.2021.3138527}
}

@article{vizzo2023kiss,
  author   = {Vizzo, Ignacio and Guadagnino, Tiziano and Mersch, Benedikt and Wiesmann, Louis and Behley, Jens and Stachniss, Cyrill},
  journal  = {IEEE Robotics and Automation Letters},
  title    = {KISS-ICP: In Defense of Point-to-Point ICP – Simple, Accurate, and Robust Registration If Done the Right Way},
  year     = {2023},
  volume   = {8},
  number   = {2},
  pages    = {1029-1036},
  doi      = {10.1109/LRA.2023.3236571}
}

@article{xu2022fastlio2,
  author   = {Xu, Wei and Cai, Yixi and He, Dongjiao and Lin, Jiarong and Zhang, Fu},
  journal  = {IEEE Transactions on Robotics},
  title    = {FAST-LIO2: Fast Direct LiDAR-Inertial Odometry},
  year     = {2022},
  volume   = {38},
  number   = {4},
  pages    = {2053-2073},
  doi      = {10.1109/TRO.2022.3141876}
}

@article{xu2021fastlio,
  author   = {Xu, Wei and Zhang, Fu},
  journal  = {IEEE Robotics and Automation Letters},
  title    = {FAST-LIO: A Fast, Robust LiDAR-Inertial Odometry Package by Tightly-Coupled Iterated Kalman Filter},
  year     = {2021},
  volume   = {6},
  number   = {2},
  pages    = {3317-3324},
  doi      = {10.1109/LRA.2021.3064227}
}

@inproceedings{lv2020targetless,
  author    = {Lv, Jiajun and Xu, Jinhong and Hu, Kewei and Liu, Yong and Zuo, Xingxing},
  booktitle = {2020 IEEE/RSJ International Conference on Intelligent Robots and Systems (IROS)},
  title     = {Targetless Calibration of LiDAR-IMU System Based on Continuous-time Batch Estimation},
  year      = {2020},
  volume    = {},
  number    = {},
  pages     = {9968-9975},
  doi       = {10.1109/IROS45743.2020.9341405}
}

@article{li2021structured,
  author   = {Li, Shuaixin and Wang, Li and Li, Jiuren and Tian, Bin and Chen, Long and Li, Guangyun},
  journal  = {IEEE Access},
  title    = {3D LiDAR/IMU Calibration Based on Continuous-Time Trajectory Estimation in Structured Environments},
  year     = {2021},
  volume   = {9},
  number   = {},
  pages    = {138803-138816},
  doi      = {10.1109/ACCESS.2021.3114618}
}

@article{li2023gnss,
  author   = {Li, Shengyu and Li, Xingxing and Zhou, Yuxuan and Xia, Chunxi},
  journal  = {IEEE Transactions on Instrumentation and Measurement},
  title    = {Targetless Extrinsic Calibration of LiDAR–IMU System Using Raw GNSS Observations for Vehicle Applications},
  year     = {2023},
  volume   = {72},
  number   = {},
  pages    = {1-11},
  doi      = {10.1109/TIM.2023.3267527}
}

@book{barfoot2024state,
  title={State estimation for robotics},
  author={Barfoot, Timothy D},
  year={2024},
  publisher={Cambridge University Press}
}

@dataset{anonymous_2026,
  author       = {Anonymous, A.},
  title        = {{LiDAR-IMU Data for Sensor Calibration on Tilted Ground}},
  year         = {2026},
  publisher    = {Zenodo},
  doi          = {10.5281/zenodo.22098085},
  url          = {https://doi.org/10.5281/zenodo.22098085}
}
